\documentclass[pmlr,twocolumn,10pt]{jmlr}

\mlhtrack{proceedings}
\jmlrproceedings{}{Submitted to ML4H 2026: \mlhtrackname}
\jmlrworkshop{Machine Learning for Health (ML4H) 2026}
\usepackage{float}
\usepackage{booktabs}
\usepackage[switch]{lineno}
\usepackage{mdframed}
\usepackage[most]{tcolorbox}
\usepackage{listings}
\usepackage{array}
\usepackage{ragged2e}

\definecolor{rowcollight}{gray}{0.95}
\definecolor{correctgreen}{HTML}{0B7A53}
\definecolor{errorred}{HTML}{B42318}
\newcommand{\okpred}[1]{\textcolor{correctgreen}{\textbf{\checkmark}}~#1}
\newcommand{\badpred}[1]{\textcolor{errorred}{\textbf{X}}~#1}
\newcolumntype{P}[1]{>{\raggedright\arraybackslash}p{#1}}
\newcommand{\oneattr}[1]{\texttt{#1}}
\newcommand{\twoattr}[2]{\begin{tabular}[t]{@{}l@{}}\texttt{#1}\\\texttt{#2}\end{tabular}}

\tcbuselibrary{listings,breakable}
\tcbset{
  umaListing/.style={
    enhanced,
    breakable,
    colback=gray!4,
    colframe=gray!55,
    boxrule=0.45pt,
    arc=1pt,
    left=2mm,
    right=2mm,
    top=1mm,
    bottom=1mm,
    fonttitle=\bfseries\scriptsize,
    coltitle=black,
    colbacktitle=gray!12,
    toptitle=0.6mm,
    bottomtitle=0.6mm,
    listing only,
    listing options={
      basicstyle=\ttfamily\tiny,
      breaklines=true,
      columns=fullflexible,
      keepspaces=true,
      showstringspaces=false
    }
  }
}

\lstdefinelanguage{json}{
    basicstyle=\ttfamily\small,
    numbers=left,
    numberstyle=\tiny,
    stepnumber=1,
    numbersep=5pt,
    showstringspaces=false,
    breaklines=true,
    frame=lines,
    backgroundcolor=\color{gray!10},
    literate=
     *{0}{{{\color{blue}0}}}{1}
      {1}{{{\color{blue}1}}}{1}
      {2}{{{\color{blue}2}}}{1}
      {3}{{{\color{blue}3}}}{1}
      {4}{{{\color{blue}4}}}{1}
      {5}{{{\color{blue}5}}}{1}
      {6}{{{\color{blue}6}}}{1}
      {7}{{{\color{blue}7}}}{1}
      {8}{{{\color{blue}8}}}{1}
      {9}{{{\color{blue}9}}}{1}
      {:}{{{\color{red}:}}}{1}
      {,}{{{\color{red},}}}{1}
      {\{}{{{\color{orange}\{}}}{1}
      {\}}{{{\color{orange}\}}}}{1}
      {[}{{{\color{orange}[}}}{1}
      {]}{{{\color{orange}]}}}{1}
      {"}{{{\color{green}"}}}{1}
}

\definecolor{graybg}{HTML}{F5F5F5}
\newcommand\boxedname{Prompt\xspace}
\newcounter{prompt}
\newenvironment{prompt}[1][]{
    \refstepcounter{prompt}
    \definecolor{graybg}{HTML}{f1f1f1}
    \begin{mdframed}[
        innertopmargin=2pt,
        innerbottommargin=2pt,
        innerleftmargin=2pt,
        innerrightmargin=2pt,
        frametitle={\textbf{\boxedname \theprompt:} #1},
        frametitlefont=\scriptsize,
        frametitlerule=true,
        backgroundcolor=graybg]%
    \setlength{\parindent}{0pt}%
    \setlength{\parskip}{0em}%
    \scriptsize\ttfamily\hyphenchar\font=`\-\spaceskip=.5em plus .5em\xspaceskip=.5em%
}
{%
    \par%
    \end{mdframed}%
}

\title[From Analytics to Tumor Boards]{From Analytics to Tumor Boards: An Evidence-Linked Multi-Agent Workflow for Oncology Feature Extraction}
\author{\Name{Daniel Kang} \\ \addr Nimblemind
\AND
\Name{Michelle Hu} \\ \addr Nimblemind
\AND
\Name{Soorya Ram Shimgekar} \\ \addr Nimblemind
\AND
\Name{Shayan Vassef} \\ \addr Nimblemind
\AND
\Name{Yufan Wang} \\ \addr Nimblemind
\AND
\Name{Anit Kumar Sahu} \\
\addr Independent Researcher
\AND
\Name{Munmun De Choudhury} \\
\addr School of Interactive Computing, Georgia Institute of Technology
\AND
\Name{Vedant Das Swain} \\
\addr Department of Technology Management and Innovation, NYU Tandon School of Engineering, New York University
\AND
\Name{Christian Poellabauer} \\
\addr Florida International University
\AND
\Name{Li Yan Khor} \\
\addr Duke-NUS Medical School and Department of Anatomical Pathology, Singapore General Hospital
\AND
\Name{Koustuv Saha} \\
\addr University of Illinois Urbana-Champaign
\AND
\Name{Robert Wojciechowski} \\
\addr OncoLens
\AND
\Name{Elliot Kidd} \\
\addr OncoLens 
\AND
\Name{Piyum Zonooz} \\ \addr Nimblemind
\AND
\Name{Navin Kumar} \\ \addr Nimblemind
}

\begin{document}
\maketitle

\ifmlhdemo\else
\begin{abstract}
Clinically relevant oncology information is distributed across heterogeneous, longitudinal documentation, creating substantial abstraction burden and requiring accurate attribution across specimens, tumors, biomarkers, and time points, while manual cancer-registry abstraction can require 27.2 minutes per case, highlighting the need for scalable methods that preserve clinical context while converting documentation into structured data. We evaluate an oncology information-extraction workflow in which OncoLens supplies multi-source, oncology-aware document selection, aggregation, and normalization from integrated EHRs, while the NimbleMind Multi-Agent System (nMAS) is a configurable oncology information-extraction workflow that extracts clinically relevant structured fields from fragmented oncology documentation. The extraction task uses a clinician-informed schema of 328 attributes spanning report metadata, diagnosis, staging, and cancer-type-specific information. nMAS separates clinician-defined field specifications from model execution and combines complexity-aware extraction, report-level consolidation, and source-grounded validation. The retrospective evaluation included 230 de-identified oncology documents from 40 patients and 418 clinician-reviewed document-field pairs containing 1,126 non-empty reference values. Evaluation focused on fields identified by clinicians as present in the source documents rather than exhaustively annotating all 328 schema fields.
nMAS achieved a rank-weighted value-level precision of 82.6\%, recall of 87.5\%, and F1 of 85.0\%, compared with an F1 of 66.4\% for an independently implemented UMA-style MiniMax M2.5 comparator. These findings support the feasibility of using a configurable, source-grounded extraction workflow to convert fragmented oncology documentation into reusable structured data. 

\end{abstract}

\begin{keywords}
Multi-Agent Systems, Medical Data Processing, Feature Identification and Enrichment, Data Extraction/Retrieval, Machine Learning Pipeline Optimization
\end{keywords}
\fi

\ifmlhneedsstatements




\section{Introduction}

Electronic health records contain large volumes of clinically relevant information, but locating and transforming that information into structured, reusable data remains a substantial challenge across healthcare.
In a previous study, physicians spent 49.2\% of the office day on electronic health record (EHR) and desk work \citep{sinsky2016physiciantime}; in a large integrated health system, more than 12 million EHR search activities were recorded over approximately 13 months, and 75.6\% of active physician users used the EHR search function \citep{ruppel2020ehrsearch}.
This information-retrieval challenge is particularly pronounced in oncology, where clinically important variables are distributed across heterogeneous documents and longitudinal episodes of care.
A scoping review of 123 cancer NLP studies found that 47\% used pathology reports, 35\% used clinical notes, and 33\% used radiology reports, illustrating the heterogeneity of source documents used in cancer information extraction \citep{wang2022cancerehr}.
In practice, oncology abstraction may require integrating pathology reports, radiology reports, cytology reports, treatment histories, and clinical notes while preserving the context in which each finding was documented \citep{dahl2025performance,gholipour2023review}.
Relevant variables may describe tumor characteristics, staging, biomarkers, treatment context, disease progression, recurrence, and metastatic disease, and may be repeated, nested, multivalued, or temporally qualified.
A single record may also contain multiple specimens, lesions, tumors, historical findings, or biomarker entries, requiring an extraction system not only to identify a clinical concept but also to associate it with the correct clinical object, time point, and disease context.
These characteristics make reliable oncology abstraction a problem of attribution, contextual interpretation, and structured representation rather than entity recognition alone.

Existing clinical information-extraction approaches address parts of this problem but do not fully satisfy these requirements.
Traditional systems are often developed for predefined variables, document types, or clinical domains and may require additional annotation, rules, or model adaptation when the requested extraction schema changes \citep{wang2018clinical,gao2018han}.
Large language models provide greater flexibility for extracting newly specified attributes, but flexible prompting alone does not guarantee consistent handling of repeated entities, multivalued fields, temporal context, schema granularity, or supporting source evidence \citep{truhn2024gpt4path,wong2025uma}.
A configurable oncology extraction workflow must therefore accommodate changing field definitions while maintaining consistent report-level outputs and traceability to the underlying documentation.

To address this need, we evaluate the feature-extraction pathway of the Nimblemind Multi-Agent System (nMAS), a configurable framework for extracting structured oncology variables from heterogeneous clinical documents. The evaluated workflow uses a clinician-informed 328-field schema to identify and consolidate clinical information across reports, with extraction complexity determining the appropriate processing pathway and source-grounded validation linking outputs to supporting documentation. This schema can be extended without retraining models or rewriting extraction code. The tiered design matches extraction strategy to field complexity, combining deterministic pattern-based extraction for highly structured fields with increasingly contextual language-model-based extraction for more complex clinical information. Clinical context is preserved by accounting for document structure, repeated findings, and attribution across specimens, tumors, and clinical time points, such as distinguishing findings from separate specimens or historical from current disease information. We evaluate nMAS against an independently implemented UMA-style Mini-Max M2.5 comparator using clinician-reviewed document-field pairs and reference values. By converting distributed oncology information into reusable, source-linked structured variables, this workflow may reduce manual information-retrieval and data-assembly effort and support registry, quality-improvement, research, and multidisciplinary case-preparation workflows~\citep{hammer2020tumorboard}.

\section{Related Work}

\noindent\textbf{Clinical Data Abstraction and Oncology-Specific Burden.}
Across healthcare, clinical data abstraction is used for administrative coding, quality improvement, clinical registries, cohort construction, and research.
Depending on the target variables and available data, abstraction may be performed through manual chart review, structured database queries, or natural language processing (NLP)-based extraction \citep{alzubi2021ehrabstraction,wang2018clinical}.
Structured queries and automated extraction can reduce manual effort when relevant variables are represented consistently in discrete EHR fields, whereas narrative or context-dependent information often requires review or more sophisticated text processing.

Oncology intensifies these general abstraction challenges because clinically relevant information is distributed across heterogeneous and longitudinal documentation and may require integration of pathology, radiology, cytology, treatment, and clinical-note evidence \citep{gholipour2023review,dahl2025performance}.
In one hospital-based cancer-registry evaluation, mean manual abstraction time was 27.17 minutes per case and decreased to 15.07 minutes with an EMR-integrated abstraction system \citep{oak2026oncoinsight}.
A related information-assembly burden occurs in multidisciplinary tumor boards, where clinicians must reconcile pathology, radiology, staging, treatment, and longitudinal disease information before case discussion.
A prospective study found that a digital tumor-board preparation platform reduced average preparation time by 23\% \citep{hammer2020tumorboard}.
These studies demonstrate that automation can reduce information-assembly burden, while also highlighting the difficulty of oncology abstraction when clinically important variables remain distributed across heterogeneous source documents.

\noindent\textbf{Approaches to Oncology Information Extraction.}
Existing clinical information-extraction approaches can address important subsets of the oncology abstraction problem.
Rule-based systems can encode terminology, report sections, negation, regular expressions, and normalization rules and are well suited to fields with stable lexical, coded, or numeric evidence \citep{wang2018clinical}, but require manual maintenance as terminology, report structure, institutional conventions, or requested fields change.

Supervised and neural approaches provide contextual modeling for predefined clinical attributes, including hierarchical pathology structure and cancer attributes across report types and disease contexts \citep{gao2018han,gao2019hisan,park2021transfer,yoon2024fastmpn}. However, oncology information-extraction systems remain commonly tailored to specific cancer sites, document types, and predefined variables, limiting adaptation to new biomarkers, cancer-specific attributes, treatment context, and nested or repeated entities without additional annotation, retraining, or rule development \citep{gholipour2023review,dahl2025performance}. Large language models offer greater flexibility through zero- and few-shot structured abstraction, with recent work evaluating proprietary and open-source models across histopathology and oncology tasks \citep{truhn2024gpt4path,balasubramanian2025mrie,grothey2025benchmark,lu2026balancing}. Schema-conditioned approaches such as Universal Abstraction (UMA) further support configurable field-level extraction without task-specific models \citep{wong2025uma}, while modular and agentic systems extend this paradigm through retrieval, field-specific extraction, validation, aggregation, and source-linked review \citep{gupta2025harmone,aalabdulsalam2026multiagent}. Together, these approaches demonstrate the potential of flexible LLM-based oncology abstraction, but a key challenge remains: adapting a single workflow to heterogeneous oncology fields with substantially different lexical, contextual, and cancer-specific reasoning requirements.

The motivation for nMAS is therefore not that existing approaches cannot be used in this setting, but that different oncology fields exhibit different evidence patterns and may benefit from different extraction mechanisms within the same configurable workflow.
nMAS combines clinician-defined field specifications with complexity-aware routing across NER/NLP-, smaller-language-model-, and larger-language-model-based extraction, followed by report-level consolidation and source-grounded validation.
This design is intended to support heterogeneous and changing oncology schemas while allowing highly regular fields and context-dependent fields to be processed through different extraction strategies within a common output and validation framework.

\section{Data}

\noindent\textbf{Dataset Overview.}
The retrospective evaluation dataset consists of 230 de-identified oncology documents from 40 patients, including pathology reports, radiology summaries, cytology reports, and oncology clinical notes. Data was from an academic medical center in REDACTED. 
All documents were de-identified before analysis, with direct patient identifiers removed or replaced by structured placeholders while preserving clinically relevant wording, section structure, diagnostic terminology, and contextual cues needed for extraction.
Each document-level record contains the source report text and, where available, associated metadata such as document identifier, de-identified patient identifier, report type, oncology context, and review-workbook annotations.
The reports span heterogeneous oncology documentation styles, including semi-structured synoptic pathology reports, free-text progress notes, imaging impressions, cytology narratives, and mixed-format clinical summaries.
This variability introduces extraction challenges including long-context reasoning, section-specific interpretation, repeated specimens, negated findings, cancer-type-specific terminology, and normalization of numeric, categorical, date, and redacted-placeholder values.
A representative de-identified oncology document excerpt and its corresponding structured field mappings are provided in \S\ref{data_examples}.

\noindent\textbf{Clinician-Informed Field Schema and Evaluation Surface.}
The extraction task is defined by a clinician-informed oncology field schema of 328 attributes spanning report metadata, diagnosis, staging, biomarkers, and cancer-type-specific fields, each with an expected value type, clinical description, and extraction guidance. Reference values were derived from clinician review workbooks, restricted to the reviewed, non-empty document-field pairs, yielding 418 pairs and 1,126 non-empty reference values after multivalue expansion. Evaluated fields were also assigned three clinician-defined importance ranks used for rank-weighted evaluation, consistent with established practice in oncology data standards~\citep{torous2021capprotocols,osterman2020mcode} and with recent work weighting extraction evaluation metrics by field importance~\citep{zhao2025emrmodel}. Full schema examples, evaluation-surface construction, and rank-weighting detail are provided in \S\ref{app:schema_evaluation_detail}.

\section{Method}
\label{sec:method}

\begin{figure}[!t]
    \centering
    \includegraphics[width=\columnwidth]{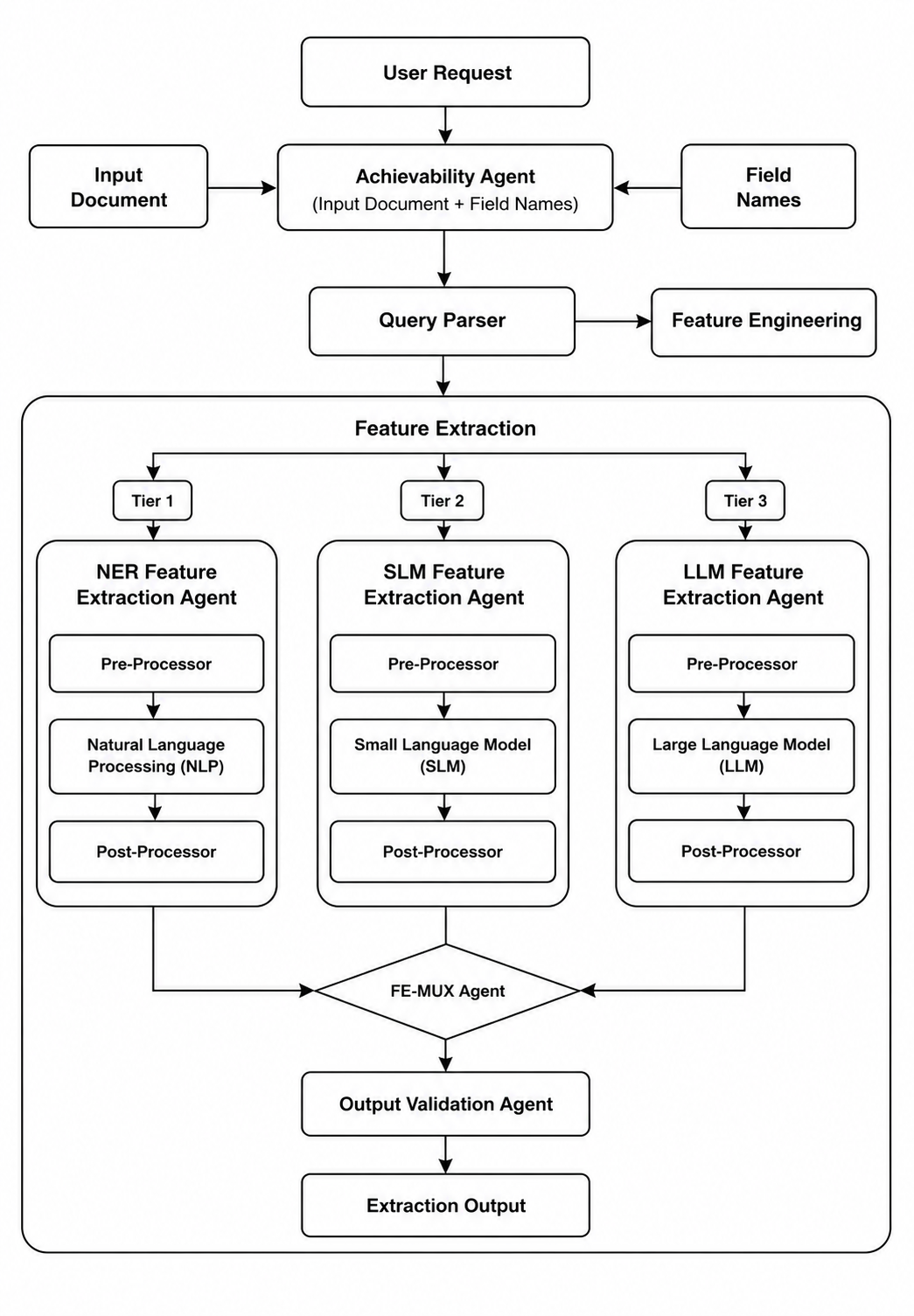}
    \caption{nMAS feature-extraction workflow. Requested fields are routed to NER-, SLM-, or LLM-based tiers, consolidated by FE-MUX, and source-validated before report-level output.}
    \label{fig:nmas}
\end{figure}

\noindent\textbf{Workflow Overview.}
As shown in Fig.~\ref{fig:nmas}, nMAS supports multiple clinical data-processing pathways \citep{wang2026trustverify}, including feature engineering and feature extraction, only the feature-extraction pathway is evaluated in this study. 

The feature-extraction pathway contains three complementary tiers. Fields with highly regular textual or coded evidence are routed to the NER/NLP-based tier, fields requiring bounded clinical interpretation to the smaller-language-model tier, and fields requiring broader contextual reasoning to the larger-language-model tier.
The resulting field-level outputs are consolidated by the feature extraction multiplexer (FE-MUX) Agent and then checked by the Output Validation Agent for schema compliance and source grounding before the final structured output is returned.

\noindent\textbf{Input Standardization and Field Preparation.}
Before extraction, each report was represented as de-identified clinical free text in a consistent textual format. Basic formatting variation, including line breaks, spacing, section separators, and repeated administrative headers, was normalized where appropriate. Clinically relevant content was preserved, including diagnostic text, synoptic pathology elements, imaging findings and impressions, staging statements, biomarker and ancillary-test results, specimen descriptions, treatment context, metastatic disease descriptions, negation, and structured redaction placeholders.

Each requested field was linked to an entry in the configurable oncology-specific \texttt{FIELD\_LIBRARY}. Each entry defined the canonical field name, expected value type or structure, clinical description, example values, relevant document context, field-specific guardrails, and, where applicable, cancer-type-specific interpretation rules.
General oncology fields applied across cancer types, whereas disease-specific fields were organized under breast, lung, ovarian, prostate, colorectal, lymphoma, and neuroendocrine categories. These field definitions guided prompt construction, extraction routing, output normalization, and source-grounded validation.

Table~\ref{tab:field_library_examples} shows representative field-library entries illustrating how target fields were paired with expected output formats and clinically informed extraction guidance.

\noindent\textbf{Achievability Agent and Query Parsing.}
The Achievability Agent assessed whether the submitted inputs contained usable oncology document content and whether the requested task corresponded to a supported feature-extraction workflow. Requests with missing inputs, unreadable files, or unsupported extraction objectives were flagged before downstream model calls were made. This step reduced the risk of producing unsupported values when the required input evidence or extraction objective was absent.

For supported extraction requests, the Query Parser resolved requested target field names against the configured oncology field library.
Matched fields were mapped to their canonical schema names and associated metadata, including expected value type, clinical description, example values, category, and cancer-type-specific context where applicable. The parser also recorded unmatched or unresolved requested fields so that schema mismatches could be reviewed explicitly rather than silently ignored. Reference labels and review-workbook values were not used during request assessment or query parsing.

\noindent\textbf{Section-Aware Context Construction.}
After query parsing, each report was segmented into common oncology document regions, including metadata, diagnosis, synoptic report, gross description, microscopic description, findings, impression, addendum, and comments. Conservative section-matching rules were used to avoid treating ordinary clinical sentences as section headings, and unmatched text was retained so that potentially relevant evidence was not discarded. Section and sentence boundaries were then used to localize supporting evidence and construct field-relevant context windows.
For fields requiring broader interpretation, the extraction agents retained access to the complete source report rather than relying exclusively on the localized section.

\noindent\textbf{Tiered Feature Extraction and Prompt Design.}\label{sec:tiered_extraction} After query parsing and section-aware context construction, each supported field request was assigned to one of three extraction routes according to its expected evidence pattern and contextual requirements.
Tier~1, the NER/NLP-based extraction tier, was used for fields supported primarily by direct lexical, coded, or highly regular textual evidence.
This route included preprocessing, natural language processing or pattern-based extraction, and postprocessing to normalize the extracted value into the required field format.

Tier~2, the Small Language Model extraction tier, was used for bounded contextual extraction.
Fields in this tier included structured or semi-structured clinical attributes for which the relevant evidence was usually localized, such as report metadata, staging context, tumor characteristics, margin status, specimen-level attributes, selected radiology findings, biomarkers, and cancer-type-specific pathology variables.
When available, section cues were used to provide the model with the most relevant report context while retaining the source text required for grounded extraction.

Tier~3, the Large Language Model extraction tier, was used for fields requiring broader contextual interpretation, repeated evidence structures, or more complex clinical synthesis.
These included fields requiring interpretation across diagnosis, synoptic report, findings, impression, addendum, or biomarker sections, as well as fields requiring careful handling of negation, uncertainty, metastatic disease, progression, or cancer-type-specific clinical logic.

Model assignments were determined before the present oncology evaluation through engineering experiments using the same tiered nMAS workflow on a separate pathology information-extraction task.
The broader candidate screening included the models ultimately selected for deployment, Qwen2.5-7B-Instruct and DeepSeek-V4-Flash, together with additional candidate models including GLM-5, Gemini~3.1~Pro, DeepSeek~V3.2, Kimi~K2~Thinking, and Qwen3-Next-80B~\citep{yang2024qwen25,deepseekai2026deepseekv4,glm5_2026,google2026gemini31pro,deepseekai2025deepseekv32,moonshotai2025kimik2thinking,qwen2025qwen3next}.
Models were compared based on field-level extraction performance, structured-output reliability, source-text grounding, negation handling, and runtime efficiency. Following this screening, Qwen2.5-7B-Instruct was selected for Tier~2 and DeepSeek-V4-Flash was selected for Tier~3. The deployed models were subsequently fine-tuned on a separate proprietary multimodal clinical dataset before integration into the nMAS extraction workflow. The 40 oncology documents used in the present evaluation and their associated reference labels were not included in the fine-tuning data and were not used for model selection or prompt development.

The language-model extraction prompt combined the selected report context with the corresponding oncology field-library entries.
These entries supplied the clinical definition, expected value type or structure, representative example values, relevant document context, and field-specific guardrails for each requested feature. Field-specific guardrails defined the extraction boundary for clinically nuanced variables, including the handling of negation, uncertainty, metastatic disease, progression, biomarker status, and cancer-type-specific pathology logic. For example, staging fields required explicit staging statements rather than inference from tumor size alone, biomarker fields were anchored to reported immunohistochemistry or molecular results, and metastatic-site fields required explicit evidence of metastatic disease while respecting negated mentions. The shared instruction required extraction only from information explicitly present in the source report, prohibited unsupported inference, and required structured JSON output with supporting source text when available. If a requested value was absent, unsupported, or not explicitly stated, the model was instructed to return \texttt{NOT\_FOUND}.

The core oncology extraction instruction and a representative field-specific guardrail are provided in Appendix~\ref{app:oncology_prompt}.

\noindent\textbf{Field-Specific In-Context Demonstrations and Guardrails.}
For the language-model extraction tiers, the oncology extraction prompts included a configured set of field-specific in-context mapping demonstrations.
Each demonstration paired a short oncology report-style excerpt or clinical pattern with the expected schema mapping and a guardrail defining the relevant extraction boundary.
The demonstrations addressed recurrent oncology extraction challenges, including diagnosis mapping, repeated cancer and biomarker indexing, tumor-size normalization, separation of staging components, biomarker result grouping, cancer-specific pathology interpretation, and collection-date extraction.

Field-specific guidance was also supplied through the oncology \texttt{FIELD\_LIBRARY}.
For each requested field, the field library provided clinical context, the expected value type or structure, representative example values, and guardrails defining the extraction boundary.
These guardrails served as prompt-level constraints designed to limit unsupported inference, preserve negation and uncertainty, and reduce common false-positive patterns.
The demonstrations and guardrails were supplied only to the language-model extraction tiers; the evaluated report-level reference values were not included in the extraction prompts.
Representative examples are shown in Appendix Table~\ref{tab:oncology_icl_examples}.

\noindent\textbf{Feature Extraction Multiplexing.}
Outputs from the active extraction tiers were passed to the Feature Extraction Multiplexer Agent (FE-MUX), which assembled field-level predictions into a single report-level structured record.
The multiplexer consolidated available NER/NLP-, SLM-, and LLM-derived outputs for the requested fields and retained associated extraction metadata where available, including supporting source text, confidence information, dates, and explanatory notes.
This step produced a unified JSON representation for each report before downstream validation. Schema compliance and source-grounding checks were performed by the subsequent validation stage.

\noindent\textbf{Output Validation and Generation.}
The Output Validation Agent assessed both output structure and source grounding before final generation. This step used deterministic string matching. For each retained field value, the reported supporting sentence was located in the original de-identified report, and the extracted value was checked for containment within that sentence. Each field was annotated with its match outcome rather than discarded, allowing downstream review to distinguish source-grounded values from those that could not be located verbatim in the report. Values marked as absent, unsupported, or \texttt{NOT\_FOUND} were not subjected to evidence matching.

Validation also checked whether extracted values conformed to the expected field structure and whether clinically important boundaries were preserved, including negation, uncertainty, metastatic disease status, biomarker interpretation, and progression-related language.
Missing, unsupported, or internally inconsistent outputs were flagged for review rather than accepted as fully grounded predictions.
The generation stage then assembled the validated fields into structured report-level JSON outputs and reviewable audit files for downstream evaluation.

\noindent\textbf{External UMA-Style Benchmark.}
We implemented an external comparator modeled after Universal Abstraction (UMA)~\citep{wong2025uma}, using one-attribute prompting to extract each target field independently. The comparator was evaluated on the same reviewed document-field pairs as nMAS, with reference values and nMAS outputs excluded from the prompts to prevent leakage. The final benchmark used MiniMax M2.5, and its outputs were normalized to the same evaluation format as nMAS before scoring. Additional details on model selection, prompting, and audit examples are provided in \S\ref{app:uma_details}.

\noindent\textbf{Evaluation Metrics.}\label{sec:evaluation_metrics}
Gold-standard labels underwent external clinician review against source documents before evaluation. Value-level precision, recall, and F1 were computed over matched, unmatched predicted, and unmatched reference values (TP/FP/FN) within the reviewed, non-empty evaluation surface described above, and two complementary document-field recovery scores (strict and lenient) were also reported. Rank-weighted overall scores used the prespecified 43\%/30\%/27\% weights for Ranks~1--3. Full formulas, the gold-standard reconciliation procedure, and detailed value normalization and matching procedures are provided in \S\ref{app:schema_evaluation_detail} and \S\ref{sec:appendix_evaluation}.

\noindent\textbf{Ablation Study.}\label{sec:ablation_method}
To isolate the contribution of individual pipeline components, eleven conditions were evaluated on a held-out subset of 40 documents from 6 patients, the portion of the evaluation corpus that underwent formal audit and reconciliation against source documents. The small subset size reflects the cost of multi-annotator reconciliation at this confidence level. Conditions were run in either an isolated-extraction or full-pipeline configuration and scored with the same restricted, set-based matching methodology used for the main evaluation (\S\ref{sec:evaluation_metrics}, Appendix~\ref{sec:appendix_evaluation}). Full setup details and the condition list are given in Appendix~\ref{sec:ablation_appendix_full}.

\section{Results}

\begin{table*}[!t]
\floatconts
  {tab:rank_stratified_results}%
  {\caption{Rank-stratified extraction performance for nMAS and the UMA + MiniMax M2.5 comparator.
  The rank-weighted overall score uses clinician-defined weights of 43\%, 30\%, and 27\% for Ranks~1--3, respectively.}}%
  {%
\centering
\footnotesize
\setlength{\tabcolsep}{2.6pt}
\renewcommand{\arraystretch}{1.05}

\resizebox{0.92\textwidth}{!}{%
\begin{tabular}{
@{}
>{\raggedright\arraybackslash}p{0.085\textwidth}
>{\raggedright\arraybackslash}p{0.235\textwidth}
r
rrrrr
r
@{}
}
\toprule
\textbf{Rank} &
\textbf{Example Features} &
\shortstack{\textbf{Reviewed}\\\textbf{Doc Fields}} &
\multicolumn{5}{c}{\textbf{nMAS}} &
\shortstack{\textbf{UMA + MiniMax}\\\textbf{F1}} \\
\cmidrule(lr){4-8}

& & &
\textbf{Strict} &
\textbf{Lenient} &
\textbf{Precision} &
\textbf{Recall} &
\textbf{F1} &
\\
\midrule

Rank 1 &
disease status; metastasis; treatment line &
215 &
89.7\% &
97.3\% &
82.4\% &
89.8\% &
85.9\% &
72.6\% \\

Rank 2 &
recurrence flag; tumor size; site flags &
56 &
82.1\% &
89.1\% &
78.3\% &
82.1\% &
80.2\% &
58.0\% \\

Rank 3 &
study date; modality; body region &
147 &
91.9\% &
94.4\% &
87.9\% &
89.8\% &
88.8\% &
65.7\% \\
\midrule

\textbf{Weighted overall} &
--- &
418 &
\textbf{88.0\%} &
\textbf{94.1\%} &
\textbf{82.6\%} &
\textbf{87.5\%} &
\textbf{85.0\%} &
\textbf{66.4\%} \\

\bottomrule
\end{tabular}%
}
  }
\end{table*}

Table~\ref{tab:rank_stratified_results} summarizes rank-stratified value-level extraction performance and document-field scores for nMAS and the UMA + MiniMax M2.5 comparator.

nMAS achieved rank-weighted value-level precision, recall, and F1 scores of 82.6\%, 85.0\%, and 87.5\% respectively, computed over the reviewed, non-empty evaluation surface (\S\ref{sec:evaluation_surface}) rather than the full 328-field schema.
The strongest value-level performance was observed for Rank~3, with an F1 score of 88.8\%, whereas Rank~2 had the lowest F1 score at 80.2\%.

nMAS also achieved rank-weighted strict and lenient document-field scores of 88.0\% and 94.1\%, respectively. The higher lenient score indicates that some errors involved partial recovery of multivalued fields rather than complete failure to recover a reviewed reference value. These scores measure whether the extracted value matched the clinician-reviewed reference and whether the cited sentence was found in the source document. They do not check whether the cited sentence actually says what the extracted value claims, for example whether it reflects the wrong time point or a negated finding.

Compared with the external UMA + MiniMax M2.5 baseline, nMAS achieved a higher rank-weighted F1 score (85.0\% vs.\ 66.4\%) and outperformed the comparator across all three ranks.
The largest absolute difference was observed for Rank~3, where nMAS achieved an F1 score of 88.8\% compared with 65.7\% for UMA + MiniMax.

Representative extraction outcomes are provided in Appendix Table~\ref{tab:error_examples}. Additional examples from a qualitative review of selected nMAS outputs by clinicians are provided in Appendix Table~\ref{tab:qa_examples}, including further correct and incorrect extraction patterns. Recurring errors involved repeated-entity alignment, table-like assay interpretation, schema-granularity mismatches, and temporal disambiguation. In several cases, relevant evidence was identified but the output was incomplete, over-expanded, or represented at a different level of specificity from the reference value.
Representative errors included a shared confusion of Oncotype DX component markers with the assay name, nMAS over-inclusion of a historical date, and comparator over-expansion of a single specimen label. The study-date example was one of the few cases in which the comparator outperformed nMAS.

\noindent\textbf{Ablation Study Results.}\label{sec:ablation_results}
The ablation study evaluated eleven conditions plus two non-LLM floors on the 40-document ablation subset. Full results are provided in Appendix Table~\ref{tab:ablation_results}. The full-pipeline baseline (F1 48.9\%) scored higher than the isolated baseline (F1 44.5\%), but both remained well below the main evaluation's rank-weighted F1 of 85.0\%. Among the tier-routing conditions, the unmodified production routing table scored lowest of the four, below random reassignment, inverted assignment, and forced Tier~3 routing.

\section{Discussion}

This study evaluated nMAS as a configurable extraction workflow for heterogeneous oncology documents.
Across 418 reviewed document-field pairs containing 1,126 reference values, nMAS achieved a rank-weighted value-level F1 of 85.0\%, compared with 66.4\% for the UMA-style MiniMax M2.5 comparator.

The observed extraction performance suggests that nMAS can pre-structure heterogeneous oncology information for downstream review, although the present study did not directly measure workflow efficiency.
This distinction is important because prior oncology studies have shown measurable benefits when manual information assembly is reduced: an EMR-integrated registry system decreased mean abstraction time from 27.17 to 15.07 minutes per case~\citep{oak2026oncoinsight}, while a digital tumor-board platform reduced average case-preparation time by 23\%~\citep{hammer2020tumorboard}.
These findings provide operational context for the potential value of structured, source-grounded extraction upstream of registry and multidisciplinary workflows.

The higher rank-weighted F1 observed for nMAS relative to the UMA-style comparator (85.0\% vs.\ 66.4\%) suggests that a configurable workflow may be useful when extraction requirements span fields with substantially different evidence patterns. The two systems differ in several respects beyond architecture, including model choice and fine-tuning, so this comparison reflects a system-level performance advantage rather than evidence that the multi-agent design itself drives the improvement. Rather than requiring a separate extraction pipeline for each target field, the schema-guided workflow provides a common mechanism for producing structured outputs with supporting source evidence.
Whether this technical advantage translates into reductions in abstraction time, staffing burden, or cost should be evaluated prospectively.

The ablation study (Table~\ref{tab:ablation_results}) indicates that the production tier-routing table, a static, pre-authored lookup rather than a per-document signal, did not outperform simpler alternatives tested under the same full-pipeline configuration. Extraction batch size also showed a clear capacity-dependent effect: Tier~3 (DeepSeek-V4-Flash) tolerated all-fields-in-one-call with little cost, while Tier~2 (Qwen2.5-7B-Instruct) did not, indicating that batch-size configuration should be tuned per model rather than applied uniformly across tiers.

Several limitations qualify these findings. The evaluation included only 230 documents across 40 patients from a single institutional setting and was restricted to the reviewed, non-empty evaluation surface (\S\ref{sec:evaluation_surface}) rather than an exhaustively annotated schema, limiting assessment of cross-institutional generalizability, specificity, true-negative performance, and false positives among unreviewed fields. The evaluation also measured whether a value matched the reference and whether its cited sentence was found in the source document, not whether that sentence's meaning actually supports the value, so a cited sentence sharing words with an incorrect value, such as one describing an earlier time point or a negated finding, would still pass. Assessing this evidence faithfulness, rather than presence, is an important direction for future work. nMAS and the UMA-style comparator also differed in model choice, prompting, context construction, extraction strategy, and output processing, so the observed performance difference should be interpreted at the complete-system level rather than as evidence for any individual component. Notably, nMAS's Tier~2 and Tier~3 models were fine-tuned on a separate proprietary clinical dataset while the comparator used an off-the-shelf model. Part of the observed performance gap may therefore reflect this difference in model adaptation rather than the extraction workflow design itself, and the comparison should not be read as isolating architecture from fine-tuning effects. The ablation study's single-call Tier~3 condition, which used the base, pre-fine-tuning model, provides a partial control for this difference, though prompt structure and underlying model choice still
differ from the comparator. Additional, ablation-specific limitations are discussed in \S\ref{sec:ablation_appendix_full}.

Future work should first evaluate larger multi-institutional cohorts using a prespecified extraction surface with exhaustive positive and negative annotation, together with matched-model ablations to isolate the effects of routing, field-specific guidance, FE-MUX, and validation. Methodologically, repeated specimens, tumors, lesions, biomarkers, and treatment events motivate explicit entity linking and temporal relation modeling, while table-like assays motivate stronger table- and report-structure-aware extraction. Structured decoding and schema-aware representations may further improve nested and multivalued outputs, and exact-set evaluation could better quantify complete recovery of related values. Finally, prospective studies should assess source-evidence faithfulness and determine whether extraction improvements translate into reduced abstraction time, staffing burden, cost, and clinician workload in registry and multidisciplinary oncology workflows.

\bibliography{ref}

\appendix

\setcounter{table}{0}
\renewcommand{\thetable}{A\arabic{table}}

\section{Supplementary Material}
\label{app:supplement}

This appendix provides supplementary material supporting the oncology extraction workflow and evaluation described in the main text.
It includes representative oncology field-library entries and source-to-schema mappings; details of the external UMA-style comparator, including model selection, prompting, and audit examples; representative nMAS extraction instructions and in-context demonstrations; detailed value normalization and matching procedures used for evaluation; and qualitative examples illustrating recurring extraction outcomes and error patterns for nMAS and the UMA + MiniMax M2.5 comparator.


\subsection{Field Schema, Evaluation Surface, and Metric Detail}
\label{app:schema_evaluation_detail}

\noindent\textbf{Clinician-Informed Field Schema.}
The extraction task is defined by a clinician-informed oncology field schema containing 328 attributes spanning report metadata, diagnosis, staging, tumor characteristics, biomarkers, imaging, treatment context, and cancer-type-specific fields.
Each schema entry specifies a canonical feature name, expected value type, clinical description, example values, relevant document context, and, where applicable, cancer-specific interpretation rules.
The schema supports configurable field-name-driven extraction while preserving cancer-specific guidance.
The 328 attributes represent the full set of fields supported by the schema rather than fields expected in every document; field applicability and prevalence vary across the corpus, with common metadata fields occurring frequently and cancer-type-specific or specimen-level fields occurring less often. Representative schema fields are shown in Table~\ref{tab:feature_schema}.

\noindent\textbf{Reference Standard and Evaluation Surface.}\label{sec:evaluation_surface}
Reference values were derived from clinician review workbooks associated with the de-identified oncology documents.
Clinical reviewers were provided with the source documents and full 328-field schema and manually recorded values for fields identified as present in each document, independently of nMAS and the external comparator. Fields for which no value was recorded were not treated as confirmed negatives, as the absence of an entry does not distinguish between a field being absent from the document and not being identified during review. Accordingly, blank workbook cells were treated as unreviewed, and the primary evaluation was restricted to document-field pairs with non-empty clinician-reviewed reference  values.
Candidate document-field pairs were initially surfaced by nMAS output and then confirmed through clinician review, which reached 94\% agreement with the surfaced candidates. This evaluation surface was applied consistently to nMAS and the external baseline and contained 418 reviewed document-field pairs. Because the candidate fields originated from nMAS rather than an independent, exhaustive audit, some evaluation-surface bias may remain (see Limitations).
After repeated and list-like fields were expanded, these pairs yielded 1,126 non-empty reference values for value-level evaluation.

\noindent\textbf{Clinical Priority Metadata.}
The evaluated oncology fields were assigned to three importance ranks by senior clinicians involved in schema development, based on their relative importance to oncology practice. These clinician-defined ranks were used for rank-stratified performance analysis and for the rank-weighted overall evaluation. This approach of prioritizing clinical data elements by importance is consistent with established practice in oncology data standards~\citep{torous2021capprotocols,osterman2020mcode} and with recent work weighting extraction evaluation metrics by field importance~\citep{zhao2025emrmodel}.

\begin{table}[!t]
\floatconts
  {tab:feature_schema}%
  {\caption{Representative oncology feature targets.}}%
  {%
\scriptsize
\setlength{\tabcolsep}{3pt}
\renewcommand{\arraystretch}{1.05}
\begin{tabular}{
    >{\RaggedRight\arraybackslash}p{0.49\columnwidth}
    >{\RaggedRight\arraybackslash}p{0.19\columnwidth}
    >{\RaggedRight\arraybackslash}p{0.27\columnwidth}
}
\toprule
\textbf{Feature} & \textbf{Type} & \textbf{Example} \\
\midrule

\multicolumn{3}{l}{\textit{Imaging Context}} \\
\addlinespace[1pt]
\texttt{study\_date} & date & 2025-02-15 \\
\texttt{modality} & categorical & CT \\
\texttt{body\_region} & text & Chest/Abdomen \\

\midrule
\multicolumn{3}{l}{\textit{Disease Status}} \\
\addlinespace[1pt]
\texttt{overall\_disease\_status} & categorical & progression \\
\texttt{progression\_flag} & boolean & Y \\

\midrule
\multicolumn{3}{l}{\textit{Primary Tumor}} \\
\addlinespace[1pt]
\texttt{primary\_tumor\_site} & text & Right lung \\
\texttt{primary\_tumor\_max\_diameter\_mm} & integer & 32 \\

\midrule
\multicolumn{3}{l}{\textit{Nodal and Metastatic Disease}} \\
\addlinespace[1pt]
\texttt{nodal\_disease\_present\_flag} & boolean & Y \\
\texttt{distant\_metastasis\_present\_flag} & boolean & Y \\
\texttt{metastatic\_sites} & text & Liver; lung \\
\texttt{liver\_metastases\_present\_flag} & boolean & Y \\
\texttt{lung\_metastases\_present\_flag} & boolean & Y \\

\midrule
\multicolumn{3}{l}{\textit{Treatment Context}} \\
\addlinespace[1pt]
\texttt{treatment\_context} & categorical & \texttt{on\_treatment} \\
\texttt{inferred\_line\_of\_therapy} & integer & 2 \\
\texttt{recurrence\_suspected\_flag} & boolean & Y \\

\bottomrule
\end{tabular}
  }
\end{table}

\noindent\textbf{Evaluation Metrics Detail.}
Before model evaluation, the initial gold-standard labels underwent external review against the corresponding source documents, with reviewer agreement calculated as the proportion of reviewed labels matching the initial labels.

Evaluation was restricted to document-field pairs with non-empty clinician-reviewed reference values. Because the review workbooks did not exhaustively annotate the 328-field schema, blank cells were treated as unreviewed rather than confirmed negatives; predictions outside the reviewed evaluation surface were excluded, and true negatives were not reported. For each evaluated pair $i$, let $R_i$ and $P_i$ denote the reference and predicted value sets, respectively.

Value-level performance was evaluated using precision, recall, and F1, with matched, unmatched predicted, and unmatched reference values counted as true positives (TP), false positives (FP), and false negatives (FN), respectively. Missing or invalid predictions were treated as empty predicted sets. Detailed value normalization and matching procedures are provided in \S\ref{sec:appendix_evaluation}.

Two complementary document-field recovery scores were also reported. Let $M_i$ denote the subset of reference values successfully matched by the model. The strict score measured the proportion of reference values recovered,
$$
S^{\mathrm{strict}}_i=\frac{|M_i|}{|R_i|},
$$
while the lenient score assigned full credit when at least one reference value was recovered:
$$
S^{\mathrm{lenient}}_i=
\begin{cases}
1, & |M_i|>0,\\
0, & |M_i|=0.
\end{cases}
$$
Both scores included all reviewed document-field pairs, including those without a valid prediction; extra predicted values were penalized through value-level precision and F1.

For each clinician-assigned importance rank, TP, FP, and FN counts were pooled before calculating value-level metrics, while strict and lenient scores were averaged across document-field pairs. Rank-weighted metrics used prespecified weights of 43\%, 30\%, and 27\% for Ranks~1, 2, and 3, respectively:
$$
S_{\mathrm{weighted}}=0.43S_1+0.30S_2+0.27S_3.
$$


\subsection{Representative Oncology Field-Library Entries}
\label{app:field_library}

Table~\ref{tab:field_library_examples} provides representative entries from the oncology-specific field library used to define expected outputs and clinically informed extraction boundaries.

\begin{table}[!t]
\floatconts
  {tab:field_library_examples}%
  {\caption{Representative entries from the oncology-specific field library.}}%
  {%
\scriptsize
\setlength{\tabcolsep}{3pt}
\renewcommand{\arraystretch}{1.10}

\begin{tabular}{
    >{\RaggedRight\arraybackslash}p{0.25\columnwidth}
    >{\RaggedRight\arraybackslash}p{0.25\columnwidth}
    >{\RaggedRight\arraybackslash}p{0.45\columnwidth}
}
\toprule
\textbf{Field} &
\textbf{Expected output} &
\textbf{Clinical guidance} \\
\midrule

\path{pathologic_stage}
&
TNM or stage group
&
Use explicit staging statements; do not infer stage from tumor size alone.
\\

\path{histology}
&
Tumor type
&
Extract from diagnosis or synoptic sections; preserve the clinically specific subtype when stated.
\\

\path{margin_status}
&
Positive, negative, close, or not reported
&
Use surgical pathology margin statements; do not infer margin status from specimen orientation alone.
\\

\path{er_status}
&
Positive, negative, equivocal, or not reported
&
Use biomarker or immunohistochemistry results and retain supporting source evidence when available.
\\

\path{metastatic_site}
&
Anatomic site or not reported
&
Require explicit evidence of metastatic disease and preserve negated findings.
\\

\path{specimen_type}
&
Documented specimen type, e.g., cytology, fine-needle aspiration, core biopsy, excision, or resection
&
Record the specific specimen type as documented, including cytology and fine-needle aspiration specimens.
\\

\bottomrule
\end{tabular}
  }
\end{table}


\subsection{Representative Oncology Data Example}
\label{data_examples}

The following excerpt illustrates the type of heterogeneous oncology documentation processed in the study.
Patient-identifying information is removed or represented using de-identified placeholders.

\begin{quote}
\small
CT chest/abdomen/pelvis demonstrates interval increase in metastatic hepatic lesions, with the largest lesion measuring 32~mm, consistent with disease progression while the patient is receiving second-line therapy.
\end{quote}

The excerpt supports several structured oncology variables, as illustrated in Table~\ref{tab:data_example}.

\begin{table}[!t]
\centering
\small
\begin{tabular}{
p{0.48\columnwidth}
p{0.38\columnwidth}
}
\toprule
\textbf{Target Field} & \textbf{Example Value} \\
\midrule
\texttt{modality} & CT \\
\texttt{body\_region} & Chest/Abdomen/Pelvis \\
\texttt{overall\_disease\_status} & progression \\
\texttt{metastatic\_sites} & Liver \\
\texttt{inferred\_line\_of\_therapy} & 2 \\
\bottomrule
\end{tabular}
\caption{Representative mapping from oncology source text to structured target fields.}
\label{tab:data_example}
\end{table}


\subsection{External UMA-Style Benchmark Details}
\label{app:uma_details}

\noindent\textbf{Evaluation Setup.}
The external comparator was modeled after Universal Abstraction (UMA)~\citep{wong2025uma} using a one-attribute prompting strategy.
For each evaluated document-field pair, the model received the de-identified oncology report, one target field name, a concise field definition, the expected value type, representative examples or guardrails where available, and a compact JSON output schema.
The model was instructed to return the extracted value with supporting source evidence or an empty value set when the target attribute was absent or not explicitly supported by the report.

The comparator was evaluated on the same reviewed document-field pairs used for nMAS evaluation.
The review workbook was used only to define the evaluation surface; clinician-reviewed reference values, nMAS outputs, and correctness indicators were excluded from comparator prompts.
Comparator outputs were normalized into the same value-level evaluation format as nMAS outputs before scoring.

\noindent\textbf{Model Selection.}
The final comparator used MiniMax M2.5 on Amazon Bedrock~\citep{awsbedrock2026minimax,awsbedrock2026models}.
Model selection was based on a five-document challenging-set pilot using the same one-attribute prompting format.
Candidate models were compared on extraction performance, valid structured-output generation, source-evidence grounding, and runtime characteristics.
Table~\ref{tab:model_sweep} summarizes the pilot results.

\begin{table*}[!t]
\floatconts
  {tab:model_sweep}%
  {\caption{Five-document challenging-set pilot used to select the external UMA model. Bedrock and Azure model availability is documented by the respective provider model catalogs \citep{awsbedrock2026models,azureopenai2026models}. Latency is reported as seconds per attribute call; throughput is reported as attribute calls per minute and documents per hour.}}%
  {%
\scriptsize

\resizebox{\textwidth}{!}{
\begin{tabular}{rllrrrrrrrrrr}
\toprule
\textbf{Rank} &
\textbf{Model} &
\textbf{Provider} &
\textbf{Valid/Err} &
\textbf{Attr/min} &
\textbf{Latency (s/attr)} &
\textbf{Docs/hr} &
\textbf{P} &
\textbf{R} &
\textbf{F1} &
\textbf{TP} &
\textbf{FP} &
\textbf{FN} \\
\midrule

1 & Claude Opus 4.7 & Bedrock & 191/0 & 3.60 & 16.66 & 5.66 & 0.566 & 0.539 & 0.552 & 103 & 79 & 88 \\
2 & Claude Opus 4.8 & Bedrock & 191/0 & 2.48 & 24.15 & 3.90 & 0.564 & 0.508 & 0.534 & 97 & 75 & 94 \\
3 & MiniMax M2.5 & Bedrock & 191/0 & 8.32 & 7.21 & 13.07 & 0.564 & 0.508 & 0.534 & 97 & 75 & 94 \\
4 & Llama 4 Maverick 17B & Bedrock & 191/0 & 4.96 & 12.11 & 7.78 & 0.558 & 0.503 & 0.529 & 96 & 76 & 95 \\
5 & GPT-4o & Azure & 190/1 & 52.91 & 1.13 & 83.10 & 0.508 & 0.508 & 0.508 & 97 & 94 & 94 \\
6 & Nova Pro & Bedrock & 190/1 & 2.33 & 25.76 & 3.66 & 0.564 & 0.461 & 0.507 & 88 & 68 & 103 \\
7 & GLM 5 & Bedrock & 191/0 & 8.95 & 6.70 & 14.06 & 0.557 & 0.461 & 0.504 & 88 & 70 & 103 \\
8 & DeepSeek V3.2 & Bedrock & 191/0 & 6.10 & 9.84 & 9.58 & 0.587 & 0.440 & 0.503 & 84 & 59 & 107 \\
9 & Nemotron 3 Super 120B A12B & Bedrock & 191/0 & 15.19 & 3.95 & 23.86 & 0.523 & 0.482 & 0.501 & 92 & 84 & 99 \\
10 & GPT-5-mini & Azure & 647/0 & 14.72 & 4.08 & 6.82 & 0.475 & 0.503 & 0.489 & 96 & 106 & 95 \\
11 & Gemma 3 12B IT & Bedrock & 191/0 & 6.59 & 9.10 & 10.35 & 0.472 & 0.482 & 0.477 & 92 & 103 & 99 \\
12 & Qwen3 Next 80B A3B & Bedrock & 191/0 & 6.60 & 9.10 & 10.36 & 0.523 & 0.414 & 0.462 & 79 & 72 & 112 \\

\bottomrule
\end{tabular}
}
  }
\end{table*}

MiniMax M2.5 was selected because it lay on the pilot performance--runtime Pareto frontier.
It matched the Claude Opus 4.8 F1 score while processing attribute calls more than three times faster, was substantially faster than the top-F1 Claude Opus 4.7 run, and achieved higher F1 than the other faster Bedrock candidates.

\noindent\textbf{UMA Prompting.}
The same one-attribute prompt structure was used across evaluated fields, with the target-specific descriptor and guidance substituted for each requested attribute.
A representative MiniMax M2.5 prompt is shown below.

\begin{tcblisting}{umaListing,title={UMA prompt template (MiniMax M2.5)},label={lst:uma_prompt}}
System: You are a UMA-style clinical abstraction assistant. Return only compact valid JSON.

User:
Abstract exactly one target attribute from the DOCUMENT.

TARGET ATTRIBUTE
- data_point: disease_specific.lymphoma.ihc[].result
- type: string
- description: Lymphoma IHC result
- category: Lymphoma

FIELD-SPECIFIC GUIDANCE
- For lymphoma IHC, parse marker lists in phrases such as "positive for ... and negative for ...".
- For marker targets, return marker names only. For result targets, return the marker-specific positive/negative result.

Return exactly this JSON shape:
{"data_point":"disease_specific.lymphoma.ihc[].result",
 "values":[{"specifics":"disease_specific.lymphoma.ihc[].result",
            "value":"...",
            "evidence":"verbatim quote from DOCUMENT"}]}

Rules:
- If the target attribute is not explicitly supported, return {"data_point":"...","values":[]}.
- Use only the DOCUMENT CONTEXT. Do not infer, guess, or fill missing values.
- Evidence must be a short verbatim quote copied from the DOCUMENT CONTEXT.
- If multiple values are supported, return multiple objects in values.
\end{tcblisting}

Table~\ref{tab:uma_prompt_blocks} provides representative target-attribute blocks inserted into the shared one-attribute prompt template.

\begin{table*}[!t]
\floatconts
  {tab:uma_prompt_blocks}%
  {\caption{Representative target-attribute blocks inserted into the same UMA one-attribute prompt template. The examples illustrate handling of coding values, repeated specimens, report metadata, and biomarker panels.}}%
  {%
\footnotesize
\setlength{\tabcolsep}{4pt}
\renewcommand{\arraystretch}{1.14}

\begin{tabular}{
@{}
P{0.22\textwidth}
P{0.35\textwidth}
P{0.35\textwidth}
@{}
}
\toprule
\textbf{Target attribute} &
\textbf{Descriptor supplied to the model} &
\textbf{Attribute-specific guidance} \\
\midrule

\oneattr{clinical.coding.icd10}
&
ICD-10 or diagnosis-code values documented in the report. Type: string list. Example: \texttt{Z85.3}.
&
Return each explicit ICD-10 code as a separate value; preserve punctuation and ordering from the source text; do not infer codes from diagnoses.
\\

\twoattr{clinical.specimens.}{items[].label}
&
Specimen part label associated with the reviewed specimen item. Type: string. Example: \texttt{G}.
&
Return only labels that directly identify the target specimen item, not every part label visible in the report. Evidence must include the specimen letter and local specimen description.
\\

\twoattr{clinical.report.}{collectionDate}
&
Collection date or dates associated with the clinical report. Type: date/string list.
&
Extract every explicitly stated collection date in repeated report blocks; avoid accession, received, signed, or historical comparison dates unless the field label says collection.
\\

\twoattr{cancers[].clinical.}{biomarkers[].name}
&
Biomarker, receptor, or IHC marker names stated in the cancer-specific panel. Type: string list.
&
Return marker names only, not result words. Include negative markers when the panel explicitly lists them, because absence of positivity is still a reported biomarker result.
\\

\bottomrule
\end{tabular}
  }
\end{table*}

\noindent\textbf{Audit Examples.}
Raw model responses were retained together with normalized extracted values, allowing individual scoring decisions to be traced to model outputs and supporting source evidence.
Representative examples are shown in Table~\ref{tab:uma_audit_examples}.

\begin{table*}[!t]
\floatconts
  {tab:uma_audit_examples}%
  {\caption{MiniMax UMA audit examples showing raw JSON-style output, normalized final values, and scoring against reference values.}}%
  {%
\tiny
\setlength{\tabcolsep}{3pt}
\renewcommand{\arraystretch}{1.16}

\begin{tabular}{
@{}
P{0.17\textwidth}
P{0.31\textwidth}
P{0.14\textwidth}
P{0.18\textwidth}
P{0.12\textwidth}
@{}
}
\toprule
\textbf{Attribute} &
\textbf{MiniMax JSON-style response excerpt} &
\textbf{Final extracted value(s)} &
\textbf{Reference value(s)} &
\textbf{Score} \\
\midrule

\oneattr{clinical.coding.icd10}
&
\texttt{\{value: Z85.3\}}; \texttt{\{value: N83.209\}} with evidence from the same diagnosis-code line.
&
\texttt{Z85.3}; \texttt{N83.209}
&
\texttt{N83.209}; \texttt{Z85.3}
&
TP=2, FP=0, FN=0
\\

\twoattr{disease\_specific.}{lymphoma.ihc[].result}
&
\texttt{\{value: CD20 positive\}}; \texttt{\{value: CD5 positive\}}; \texttt{\{value: CD3 negative\}} with evidence from positive and negative IHC marker lists.
&
Six marker-specific result strings
&
\texttt{positive}; \texttt{negative}
&
TP=2, FP=4, FN=0
\\

\twoattr{clinical.specimens.}{items[].label}
&
\texttt{\{value: A\}} through \texttt{\{value: G\}}, with evidence showing specimen parts A through G.
&
\texttt{A--G}
&
\texttt{G}
&
TP=1, FP=6, FN=0
\\

\twoattr{cancers[].clinical.}{biomarkers[].name}
&
\texttt{\{value: ER\}}; \texttt{\{value: PR\}}; \texttt{\{value: p53\}}, with evidence from an IHC panel mentioning several additional negative markers.
&
\texttt{ER}; \texttt{PR}; \texttt{p53}
&
\texttt{ER}; \texttt{PR}; \texttt{p53}; \texttt{PAX8}; \texttt{WT1}; \texttt{OCT4}; \texttt{glypican 3}; \texttt{Napsin A}; \texttt{GATA3}; \texttt{p16}
&
TP=3, FP=0, FN=7
\\

\bottomrule
\end{tabular}
  }
\end{table*}

These examples show how raw comparator responses, normalized feature-level outputs, and TP/FP/FN scoring can be traced at the individual document-field level.
The largest MiniMax errors generally arose from over-broad repeated-value extraction, mismatch between marker-specific and reference granularity, and missed values in long table-like panels rather than unsupported value generation.

\subsection{Detailed Value Normalization and Matching}\label{sec:appendix_evaluation}

\noindent\textbf{Value Normalization.}
Predicted and reference values were normalized using field-specific transformations before comparison. Text values underwent case folding and cleanup of extraneous whitespace and punctuation. Dates were converted to a consistent representation, Boolean values were standardized to a common representation, and numeric values underwent unit-aware normalization where applicable. These transformations were applied symmetrically to model predictions and reference values to distinguish formatting differences from substantive extraction errors.

\noindent\textbf{Repeated and List-Like Fields.}
Fields capable of containing multiple values were expanded into individual value instances before comparison. This included repeated entities and list-like outputs in which multiple clinically distinct values could occur within the same document-field pair. Expansion allowed each value to be evaluated independently rather than treating a multi-value field as a single string.

\noindent\textbf{One-to-One Matching.}
Following normalization and expansion, predicted and reference values were matched one-to-one within each document-field pair. Each predicted value could match at most one reference value, and each reference value could be matched by at most one prediction. This prevented duplicate predictions from receiving multiple matches against a single reference value.

\noindent\textbf{Invalid and Missing Outputs.}
Model outputs that were missing, invalid, unsupported, or explicitly returned \texttt{NOT\_FOUND} were treated as containing no extractable values. These outputs therefore could not produce a value-level match. The same normalization, expansion, and matching procedures were applied to nMAS and the external comparator to ensure that differences in performance reflected extraction rather than evaluation processing.


\subsection{Full Ablation Study Results}\label{sec:ablation_appendix_full}
The ablation study used the base, pre-fine-tuning Tier~2 (Qwen2.5-7B-Instruct) and Tier~3 (DeepSeek-V4-Flash) models, served independently of the main evaluation infrastructure, rather than the fine-tuned models deployed in the main evaluation (\S\ref{sec:tiered_extraction}). Ablation results are therefore not directly comparable to the main-evaluation reference rows in Table~\ref{tab:ablation_results} for this reason as well.

Each condition used one of two configurations. \emph{Isolated extraction} called the Tier~2/Tier~3 extraction logic directly, without running the rest of the pipeline (document classification, achievability checking, feature-engineering multiplexing, output validation). \emph{Full pipeline} executed the complete workflow, including Tier~1 NER/NLP extraction, achievability checking, feature-engineering multiplexing, and output validation. A baseline was evaluated in both configurations to measure the effect of the rest of the pipeline. The resulting conditions are referred to as the \emph{isolated baseline} and \emph{full-pipeline baseline}. All other conditions used the configuration relevant to the component under study.

Feature-to-tier assignment in the full pipeline is determined by exact-match lookup against a static, pre-authored routing table, not a per-document or per-instance signal. Table~\ref{tab:ablation_conditions} lists the eleven ablation conditions, grouped by the mechanism each tests.

\begin{table*}[!t]
\floatconts
  {tab:ablation_conditions}%
  {\caption{Ablation conditions tested, grouped by mechanism.}}%
  {%
\centering
\footnotesize
\setlength{\tabcolsep}{4pt}
\renewcommand{\arraystretch}{1.05}
\begin{tabular}{
@{}
>{\raggedright\arraybackslash}p{0.18\textwidth}
>{\raggedright\arraybackslash}p{0.62\textwidth}
>{\raggedright\arraybackslash}p{0.12\textwidth}
@{}
}
\toprule
\textbf{Mechanism} & \textbf{Variants} & \textbf{Configuration} \\
\midrule
Non-LLM floors & GLiNER-BioMed: detected entities mapped directly to schema fields, no generative step. BioBERT extractive QA: each field reformulated as a spa
n-extraction question. & N/A \\
Context truncation & Source documents shortened to 75\%, 50\%, and 25\% of original length. & Isolated \\
Field-specification stripping & Bare: field name and type only. No in-context example: all guidance retained except the example value. & Isolated \\
Extraction batch size & Tier~2 (default 1/call): increased to 3, and to all fields in one call. Tier~3 (default 3/call): reduced to 1, and increased to all fie
lds in one call. & Isolated \\
Sampling temperature & Increased from the production default of 0.0 to 0.7, two independent full-corpus replications. & Isolated \\
Tier-routing mechanism & Random reassignment: tier sizes preserved, fields shuffled across tiers. Inverted assignment: easiest fields routed to the hardest tier and vice versa. Forced Tier~3 routing: every field sent to Tier~3, bypassing Tier~1 and Tier~2. & Full pipeline \\
\bottomrule
\end{tabular}%
  }
\end{table*}

Table~\ref{tab:ablation_results} reports precision, recall, and F1 for all eleven ablation conditions plus the two non-LLM floor conditions, evaluated on the held-out 40-document ablation subset.

\begin{table*}[!t]
\floatconts
  {tab:ablation_results}%
  {\caption{Ablation study results on the 40-document, 6-patient held-out subset (see Method, Ablation Study). The main-evaluation row is shown for orientation only and is not directly comparable to the ablation rows below it.}}%
  {%
\centering
\footnotesize
\setlength{\tabcolsep}{4pt}
\renewcommand{\arraystretch}{1.05}
\begin{tabular}{
@{}
>{\raggedright\arraybackslash}p{0.30\textwidth}
>{\raggedright\arraybackslash}p{0.14\textwidth}
rrr
@{}
}
\toprule
\textbf{Condition} & \textbf{Configuration} & \textbf{Precision} & \textbf{Recall} & \textbf{F1} \\
\midrule
\multicolumn{5}{@{}l}{\textit{Main evaluation (230-doc corpus, reference only)}} \\
nMAS (main evaluation) & Full pipeline & 82.6\% & 87.5\% & 85.0\% \\
UMA + MiniMax M2.5 (main evaluation) & N/A & N/A & N/A & 66.4\% \\
\midrule
\multicolumn{5}{@{}l}{\textit{Baseline}} \\
Isolated baseline & Isolated & 60.9\% & 35.1\% & 44.5\% \\
Full-pipeline baseline & Full pipeline & 61.8\% & 40.4\% & 48.9\% \\
\midrule
\multicolumn{5}{@{}l}{\textit{Context truncation}} \\
75\% of document length & Isolated & 62.3\% & 33.4\% & 43.5\% \\
50\% of document length & Isolated & 57.9\% & 28.7\% & 38.4\% \\
25\% of document length & Isolated & 52.3\% & 19.8\% & 28.7\% \\
\midrule
\multicolumn{5}{@{}l}{\textit{Field-specification stripping}} \\
Bare (datapoint name and type only) & Isolated & 51.4\% & 31.8\% & 39.3\% \\
No in-context example & Isolated & 61.2\% & 36.2\% & 45.5\% \\
\midrule
\multicolumn{5}{@{}l}{\textit{Extraction batch size}} \\
Tier~2, batch size 3 (default 1) & Isolated & 61.8\% & 34.8\% & 44.5\% \\
Tier~2, all fields in one call & Isolated & 59.5\% & 21.1\% & 31.1\% \\
Tier~3, batch size 1 (default 3) & Isolated & 65.1\% & 35.8\% & 46.2\% \\
Tier~3, all fields in one call & Isolated & 60.4\% & 30.5\% & 40.5\% \\
\midrule
\multicolumn{5}{@{}l}{\textit{Sampling temperature (0.7 vs.\ production default 0.0)}} \\
Replication 1 & Isolated & 63.3\% & 34.2\% & 44.4\% \\
Replication 2 & Isolated & 62.3\% & 35.2\% & 45.0\% \\
\midrule
\multicolumn{5}{@{}l}{\textit{Tier-routing mechanism}} \\
Random routing & Full pipeline & 63.8\% & 41.3\% & 50.2\% \\
Inverted routing & Full pipeline & 61.5\% & 45.4\% & 52.2\% \\
Forced Tier~3 routing & Full pipeline & 64.4\% & 55.6\% & 59.7\% \\
\midrule
\multicolumn{5}{@{}l}{\textit{Non-LLM floors}} \\
GLiNER-BioMed (NER floor) & N/A & 19.1\% & 9.5\% & 12.7\% \\
BioBERT extractive QA floor & N/A & 22.1\% & 1.7\% & 3.2\% \\
\bottomrule
\end{tabular}%
  }
\end{table*}

The full-pipeline baseline scored higher than the isolated baseline (F1 48.9\% vs.\ 44.5\%), but both remained well below the main evaluation's rank-weighted F1 of 85.0\%; the untested pipeline stages exercised only in the full-pipeline configuration are therefore not the primary explanation for that gap. Among the tier-routing conditions, the unmodified production routing table (full-pipeline baseline, F1 48.9\%) scored lowest of the four, including relative to a random tier reassignment (50.2\%), a deliberately inverted assignment (52.2\%), and forced Tier~3 routing (59.7\%).

The ablation study carries additional, separate limitations. It was conducted on a smaller 40-document, 6-patient subset of the evaluation corpus rather than the full 230-document cohort. The random and inverted tier-routing tables were each evaluated as a single draw rather than averaged over multiple resamplings, so the specific margins reported for those two conditions carry unquantified sampling variance. Neither the isolated nor full-pipeline ablation baseline closed the gap to the main evaluation's rank-weighted F1 of 85.0\%, indicating that corpus size or corpus composition likely account for a substantial share of that gap in addition to, or instead of, any individually ablated mechanism.


\subsection{Representative Oncology Extraction Prompt}
\label{app:oncology_prompt}

The following prompt illustrates the shared extraction instruction and a representative field-specific guardrail used in the language-model extraction tiers.

\begin{prompt}[Representative oncology extraction instruction]
Use only information explicitly present in the oncology report.
Do not infer missing facts.
Interpret each requested target feature using the oncology field library, including its definition, expected value type, examples, document context, and clinical guardrails.

For each requested feature, return a structured JSON entry containing the extracted value, supporting source text when available, confidence score, and brief notes.
If the requested value is absent, unsupported, or not explicitly stated in the report, return \texttt{NOT\_FOUND}.

Target feature example: \texttt{progression\_flag}

Clinical guardrail: assign a positive value only when the report explicitly states progression, interval increase, new metastatic disease, enlarging lesions, or worsening disease burden.
Do not infer progression from an isolated measurement without comparison or explicit progression language.
\end{prompt}

Representative in-context mapping demonstrations are provided in Table~\ref{tab:oncology_icl_examples}.

\begin{table*}[!t]
\floatconts
  {tab:oncology_icl_examples}%
  {\caption{Representative oncology in-context mapping demonstrations and guardrails.}}%
  {%
\scriptsize
\setlength{\tabcolsep}{4pt}
\renewcommand{\arraystretch}{1.12}

\begin{tabular}{
p{0.19\textwidth}
p{0.31\textwidth}
p{0.28\textwidth}
p{0.16\textwidth}
}
\toprule
\textbf{Extraction Pattern} &
\textbf{In-Context Example} &
\textbf{Expected Field Mapping} &
\textbf{Guardrail Demonstrated} \\
\midrule

Diagnosis mapping
&
``Invasive lobular carcinoma, grade 2.''
&
Map histology to invasive lobular carcinoma, grade to 2, and subtype to lobular when explicitly stated.
&
Use definitive diagnosis text; do not infer a histologic subtype unless explicitly stated.
\\

\midrule

Tumor size
&
``Largest tumor focus measures 0.5 cm.''
&
Map the greatest tumor dimension to 5 mm when the measurement explicitly refers to a tumor focus.
&
Do not use overall specimen size or measurements of unrelated lesions.
\\

\midrule

Biomarker grouping
&
``Estrogen receptor (ER), IHC: 95\% positive, strong intensity.''
&
Create one ER biomarker entry containing the IHC method, positive result, 95\% staining percentage, and strong intensity.
&
Keep attributes from the same biomarker mention within a single indexed biomarker entry.
\\

\midrule

HER2 result
&
``IHC HER2/neu score: 0 (negative).''
&
Map HER2/neu to the IHC method, score 0, and negative qualitative status.
&
Do not infer HER2 status when the assay result is absent or unclear.
\\

\midrule

Staging
&
``pT4N0M0G1.''
&
Map the explicitly stated pathologic components to pT4, N0, and M0, and map G1 separately to tumor grade when that field is requested.
&
Do not combine pathologic, clinical, and post-treatment staging, and do not treat grade as a TNM component.
\\

\bottomrule
\end{tabular}
  }
\end{table*}


\subsection{Representative Extraction Outcomes and Error Patterns}
\label{app:error_examples}

Table~\ref{tab:error_examples} provides representative examples of correct extraction and recurring error patterns observed for nMAS and the UMA + MiniMax M2.5 comparator.

\begin{table*}[!t]
\floatconts
  {tab:error_examples}%
  {\caption{Representative extraction outcomes and error patterns.}}%
  {%
\tiny
\setlength{\tabcolsep}{3pt}
\renewcommand{\arraystretch}{1.16}

\begin{tabular}{
@{}
P{0.105\textwidth}
P{0.145\textwidth}
P{0.20\textwidth}
P{0.115\textwidth}
P{0.145\textwidth}
P{0.20\textwidth}
@{}
}
\toprule
\textbf{Pattern} &
\textbf{Attribute} &
\textbf{Reference values and evidence} &
\textbf{nMAS behavior} &
\textbf{UMA + MiniMax behavior} &
\textbf{Observation} \\
\midrule

Both correct
&
\oneattr{clinical.coding.icd10}
&
\texttt{N83.209}; \texttt{Z85.3}; the evidence line contains codes for an ovarian cyst and a history of breast cancer.
&
\okpred{\texttt{N83.209}; \texttt{Z85.3}}
&
\okpred{\texttt{N83.209}; \texttt{Z85.3}}
&
Short coded values with explicit labels were reliably extracted by both systems after list expansion.
\\

Both correct
&
\twoattr{disease\_specific.}{lymphoma.ihc[].marker}
&
\texttt{CD20}; \texttt{CD5}; \texttt{CD23}; \texttt{CD3}; \texttt{CD10}; \texttt{cyclin D1}; the evidence describes positive and negative marker panels.
&
\okpred{all six markers}
&
\okpred{all six markers}
&
When the target field requests marker names, both systems can parse compact IHC lists from diagnostic prose.
\\

MiniMax over-expanded
&
\twoattr{clinical.specimens.}{items[].label}
&
\texttt{G}; the evidence begins ``G) ... NEEDLE CORE BIOPSY''.
&
\okpred{\texttt{G}}
&
\badpred{\texttt{A--G}}
&
MiniMax identified the relevant specimen block but treated all visible part labels as target values, producing six FPs. nMAS preserved the granularity of the reviewed field.
\\

MiniMax missed values
&
\twoattr{clinical.report.}{collectionDate}
&
Four redacted collection dates in a multi-block report.
&
\okpred{all four dates}
&
\badpred{\texttt{null}}
&
Repeated report blocks require the model to track multiple date labels. MiniMax abstained despite the presence of relevant date evidence in the report.
\\

MiniMax partial recall
&
\twoattr{cancers[].clinical.}{biomarkers[].name}
&
Ten-marker ovarian IHC panel: ER, PR, p53, PAX8, WT1, OCT4, glypican 3, Napsin A, GATA3, and p16.
&
\okpred{all ten markers}
&
\badpred{\texttt{ER}; \texttt{PR}; \texttt{p53}}
&
MiniMax extracted the most salient markers but missed seven negative or less prominent markers, resulting in seven FNs without additional FPs.
\\

MiniMax granularity mismatch
&
\twoattr{disease\_specific.}{lymphoma.ihc[].result}
&
The reference stores generic \texttt{positive} and \texttt{negative} values, whereas the evidence maps these results to individual markers.
&
\okpred{\texttt{positive}; \texttt{negative}}
&
\badpred{\texttt{CD20 positive}; \texttt{CD5 positive}; ...}
&
MiniMax returned more specific marker-level strings, which were counted as FPs under the reference value granularity. This example illustrates a scoring and schema-alignment limitation.
\\

Both incorrect
&
\twoattr{clinical.}{biomarkers[].name}
&
\texttt{The Recurrence Score (RS)} from an Oncotype DX recurrence-score table.
&
\badpred{\texttt{HER2}}
&
\badpred{\texttt{ER}; \texttt{PR}; \texttt{HER2}}
&
Both systems confused component assay genes with the assay name. Assay-level variables may require stronger table and report-type awareness.
\\

UMA correct, nMAS over-included
&
\oneattr{study\_date}
&
One redacted study date.
&
\badpred{reference date plus extra historical date}
&
\okpred{single study date}
&
In this example, one-attribute prompting produced a more conservative output than nMAS when the report contained competing historical dates.
\\

\bottomrule
\end{tabular}
  }
\end{table*}


\subsection{Additional Examples from a Qualitative Clinician Review}
\label{app:qa_examples}

Table~\ref{tab:qa_examples} provides additional representative examples from a qualitative review of selected nMAS outputs by clinicians, conducted separately from the reported evaluation. These examples are illustrative of recurring patterns and are not scored against the study's evaluation methodology.

\begin{table*}[!t]
\floatconts
  {tab:qa_examples}%
    {\caption{Additional representative extraction outcomes from a qualitative review of selected nMAS outputs by clinicians.}}%
  {%
\tiny
\setlength{\tabcolsep}{3pt}
\renewcommand{\arraystretch}{1.16}

\begin{tabular}{
@{}
P{0.09\textwidth}
P{0.15\textwidth}
P{0.27\textwidth}
P{0.16\textwidth}
P{0.24\textwidth}
@{}
}
\toprule
\textbf{Outcome} &
\textbf{Field(s)} &
\textbf{Source excerpt} &
\textbf{nMAS output} &
\textbf{Note} \\
\midrule

Correct
&
\twoattr{pTNM, hormone receptors,}{HER2, grade, nodes, LVI}
&
``TNM: pT1c, pN1a, cM0; ER Status: Positive; PR Status: Negative; HER2 Protein Overexpression (IHC): Negative: 1+; Nottingham Grade: G2; \dots; Lymphovascular
 Invasion: Not Present''
&
\okpred{pT1c, pN1a, cM0, ER positive, PR negative, HER2 negativeesent}
&
One dense semicolon-delimited block was decomposed into 11+ fieltamination, including keeping HER2 status separate from its own IHC
score.
\\

Correct
&
\oneattr{clinical.biomarkers[].qualitative}
&
``Ambry genetics indicated moderate risk mutation in RET gene, a.''
&
\okpred{RET: mutation; PMS2: VUS}
&
Two genes with two different result classifications in one senteted to separate biomarker entries.
\\

Error
&
\oneattr{clinical.specimens.items[].*}
&
Document contained two specimens: Part A (``left lower leg mass'') and Part B (``left axillary contents level 1 and 2'').
&
\badpred{Part B only, across all five specimen fields}
&
Reviewer: ``reference is made only to specimen B.'' A single-specimen assumption caused one root cause to cascade across five related fields.
\\

Error
&
\oneattr{disease\_specific.breast.her2\_finalInterpretation}
&
``EndoPredict\textregistered\ is a gene expression assay for paty-stage breast cancer.''
&
\badpred{negative}
&
Reviewer noted this sentence states assay eligibility criteria,  result. nMAS treated the criteria as the finding.
\\

Error
&
\oneattr{clinical.biomarkers[].qualitative}
&
``KI-67: 7\% OF TUMOR CELLS.''
&
\badpred{7\% OF TUMOR CELLS}
&
Reviewer: a qualitative field should hold a category (e.g., ``Low proliferation''), not the raw quantitative value.
\\

\bottomrule
\end{tabular}%
  }
\end{table*}

\end{document}